# A Recommender System based on Idiotypic Artificial Immune Networks

Steve Cayzer[1] and Uwe Aickelin[2#]

[1]Hewlett-Packard Laboratories, Filton Road, Bristol BS12 6QZ, UK, steve.cayzer@hp.com
[2]School of Computer Science, University of Nottingham, NG8 1BB, UK, uxa@cs.nott.ac.uk
[#]corresponding author

Abstract-The immune system is a complex biological system with a highly distributed, adaptive and self-organising nature. This paper presents an Artificial Immune System (AIS) that exploits some of these characteristics and is applied to the task of film recommendation by Collaborative Filtering (CF). Natural evolution and in particular the immune system have not been designed for classical optimisation. However, for this problem, we are not interested in finding a single optimum. Rather we intend to identify a sub-set of good matches on which recommendations can be based. It is our hypothesis that an AIS built on two central aspects of the biological immune system will be an ideal candidate to achieve this: Antigen-antibody interaction for matching and idiotypic antibody-antibody interaction for diversity. Computational results are presented in support of this conjecture and compared to those found by other CF techniques.

## 1 INTRODUCTION

Over the last few years, a novel computational intelligence technique, inspired by biology, has emerged: AIS. This section introduces AIS and shows how it can be used for solving computational problems. In essence, the immune system is used here as inspiration to create an unsupervised machine-learning algorithm. The immune system metaphor will be explored, involving a brief overview of the basic immunological theories that are relevant to our work. We also introduce the basic concepts of CF.

### 1.1 OVERVIEW OF THE IMMUNE SYSTEM

A detailed overview of the immune system can be found in many textbooks, for instance [19]. Briefly, the purpose of the immune system is to protect the body against infection and includes a set of mechanisms collectively termed humoral immunity. This refers to a population of circulating white blood cells called B-lymphocytes, and the antibodies they create.

The features that are particularly relevant to our research are matching, diversity and distributed control. Matching refers to the binding between antibodies and antigens. Diversity refers to the fact that, in order to achieve optimal antigen space coverage, antibody diversity must be encouraged [15]. Distributed control means that there is no central controller, rather, the immune system is governed by local interactions between cells and antibodies.

The idiotypic effect builds on the premise that antibodies can match other antibodies as well as antigens. It was first proposed by Jerne [17] and formalised into a model by Farmer et al [11]. The theory is currently debated by immunologists, with no clear consensus yet on its effects in the humoral immune system [14]. The idiotypic network hypothesis builds on the recognition that antibodies can match other antibodies as well as antigens. Hence, an antibody may be matched by other antibodies, which in turn may be matched by yet other antibodies. This activation can continue to spread through the population and potentially has much explanatory power. It could, for example, help explain how the memory of past infections is maintained. Furthermore, it could result in the suppression of similar antibodies thus encouraging diversity in the antibody pool. The idiotypic network has been formalised by a number of theoretical immunologists [21].

There are many more features of the immune system, including adaptation, immunological memory and protection against auto-immune attack. Since these are not directly relevant to this work, they will not be reviewed here.

### 1.2 OVERVIEW OF CF

In this paper, we are using an AIS as a CF technique extending earlier work ([5], [6], [7]). CF is the term for a broad range of algorithms that use similarity measures to obtain recommendations. The best-known example is probably the "people who bought this also bought" feature of the internet company Amazon [2]. However, any problem domain where users are required to rate items is amenable to CF techniques. Commercial applications are usually called recommender systems [22]. A canonical example is movie recommendation.

In traditional CF, the items to be recommended are treated as 'black boxes'. That is, your recommendations are based purely on the votes of your neighbours, and not on the content of the item. The preferences of a user, usually a set of votes on an item, comprise a user profile, and these profiles are compared in order to build a neighbourhood. The key decisions to be made are:

- Data encoding: Perhaps the most obvious representation for a user profile is a string of numbers, where the length is the number of items, and the position is the item identifier. Each number represents the 'vote' for an item. Votes are sometimes binary (e.g. Did you visit this web page?), but can also be integers in a range (say [0, 5]).

- Similarity Measure: The most common method to compare two users is a correlation-based measure like Pearson or Spearman, which gives two neighbours a matching score between -1 and 1. Vector based, e.g. cosine of the angle between vectors, and probabilistic methods are alternative approaches.

The canonical example is the k-Nearest-Neighbour algorithm, which uses a matching method to select *k* reviewers with high similarity measures. The votes from these reviewers, suitably weighted, are used to make predictions and recommendations.

Many improvements on this method are possible [13]. For example, the user profiles are usually extremely sparse because many items are not rated. This means that similarity measurements are both inefficient (the so-called 'curse of dimensionality') and difficult to calculate due to the small overlap. Default votes are sometimes used for items a user has not explicitly voted on, and these can increase the overlap size [4]. Dimensionality reduction methods, such as Single Value Decomposition, both improve efficiency and increase overlap [3]. Other pre-processing methods are often used, e.g. clustering [1]. Content-based information can be used to enhance the pure CF approach [13], [9]. Finally, the weighting of each neighbour can be adjusted by training, and there are many learning algorithms available for this [10]. All these improvements could in principle be applied to our AIS but in the interests of a clear and uncluttered comparison we have kept the CF algorithm as simple as possible.

The evaluation of a CF algorithm usually centres on its accuracy. There is a difference between prediction (given a movie, predict a given user's rating of that movie) and recommendation (given a user, suggest movies that are likely to attract a high rating). Prediction is easier to assess quantitatively but recommendation is a more natural fit to the movie domain. We present results evaluating both these behaviours.

### 1.3 USING AN AIS FOR CF

To us, the attraction of the immune system is that if an adaptive pool of antibodies can produce 'intelligent' behaviour, can we harness the power of this computation to tackle the problem of preference matching and recommendation? Thus, in the first instance we intend to build a model where known user preferences are our pool of antibodies and the new preferences to be matched is the antigen in question.

Our conjecture is that if the concentrations of those antibodies that provide a better match are allowed to increase over time, we should end up with a subset of good matches. However, we are not interested in optimising, i.e. in finding the one best match. Instead, we require a set of antibodies that are a close match but which are at the same time distinct from each other for successful recommendation. This is where we propose to harness the idiotypic effects of binding antibodies to similar antibodies to encourage diversity.

The next section presents more details of our problem and explains the AIS model we intend to use. We then describe the experimental set-up, present and review results and discuss some possibilities for future work.

## 2 ALGORITHMS

### 2.1 APPLICATION OF THE AIS TO THE EACHMOVIE TASKS

The eachmovie database [8] is a public database, which records explicit votes of users for movies. It holds 2,811,983 votes taken from 72,916 users on 1,628 films. The task is to use this data to make predictions and recommendations. In the former case, we provide an estimated vote for a previously unseen movie. In the latter case, we present a ranked list of movies that the user might like.

The basic approach of CF, is to use information from a neighbourhood to make useful predictions and recommendations. The central task we set ourselves is to identify a suitable neighbourhood. The SWAMI (Shared Wisdom through the Amalgamation of Many Interpretations) framework [12] is a publicly accessible software for CF experiments. Its central algorithm is as follows:

*Select a set of test users randomly from the database*

*FOR each test user t*

    *Reserve a vote of this user, i.e. Hide from predictor*

    *From remaining votes create a new training user t'*

    **Select neighbourhood of k reviewers based on t'**

*Use neighbourhood to predict vote*

*Compare this with actual vote and collect statistics*

*NEXT t*

The code shown in bold indicates a place where SWAMI allows an implementation-dependent choice of algorithm. We use an AIS to perform selection and prediction as below.

## 2.2 ALGORITHM CHOICES

We use the SWAMI data encoding: $User = \{\{id_1, score_1\}, \{id_2, score_2\}...\{id_n, score_n\}\}$

Where *id* corresponds to the unique identifier of the movie being rated and score to this user's score for that movie. This captures the essential features of the data available.

Eachmovie vote data links a person with a movie and assigns a score (taken from the set {0, 0.2, 0.4, 0.6, 0.8, 1.0} where 0 is the worst). User demographic information (e.g. Age and gender) is provided but this is not used in our encoding. Content information about movies (e.g. Category) is similarly not used.

## 2.3 SIMILARITY MEASURE

The Pearson measure is used to compare two users *u* and *v*:

$$r = \frac{\sum_{i=1}^{n}(u_i - \bar{u})(v_i - \bar{v})}{\sqrt{\sum_{i=1}^{n}(u_i - \bar{u})^2 \sum_{i=1}^{n}(v_i - \bar{v})^2}} \quad (1)$$

Where *u* and *v* are users, n is the number of overlapping votes (i.e. Movies for which both u and v have voted), $u_i$ is the vote of user *u* for movie *i* and $\bar{u}$ is the average vote of user *u* over all films (not just the overlapping votes). The measure is amended as follows:

$$\text{if } n = 0, \ r = NoOverlapDefault$$

$$\text{if } \sum_{i=1}^{n}(u_i - \bar{u})^2 \sum_{i=1}^{n}(v_i - \bar{v})^2 = 0, \ r = ZeroVarianceDefault \quad (2)$$

$$\text{if } n < P, \ r = \frac{n}{P}r \quad (\text{where } P = overlap\ penalty)$$

The two default values are required because it is impossible to calculate a Pearson measure in such cases. Both were set to 0. Some experimentation showed that an overlap penalty *P* was beneficial (this lowers the absolute correlation for users with only a small overlap) but that the exact value was not critical. We chose a value of 100 because this is the maximum overlap expected.

## 2.4 NEIGHBOURHOOD SELECTION

For a Simple Pearson (SP) predictor, neighbourhood selection means choosing the best k (absolute) correlation scores, where k is the neighbourhood size. Not every potential neighbour will have rated the film to be predicted. Reviewers who did not vote on the film are not added to the neighbourhood. We have chosen the SP as a benchmark for our AIS recommender because it is the de facto standard for recommender algorithms and also the usual starting point for more complex neighbourhood selection schemes. Furthermore, the AIS recommender is both sufficiently different and of similar complexity to warrant a fair comparison.

For the AIS predictor, a more involved procedure is required:

*Initialise AIS*

*Encode user for whom to make predictions as antigen Ag*

*WHILE (AIS not stabilised) & (Reviewers available) DO*

   *Add next user as an antibody Ab*

   *Calculate matching scores between Ab and Ag*

   *Calculate matching scores between Ab and other antibodies*

   *WHILE (AIS at full size) & (AIS not stable) DO*

      *Iterate AIS*

Our AIS behaves as follows: At each step (iteration) an antibody's concentration is increased by an amount dependent on its matching to the antigen and decreased by an amount which depends on its matching to other antibodies. In absence of either, an antibody's concentration will slowly decrease over time. Antibodies with a sufficiently low concentration are removed from the system, whereas antibodies with a high concentration may saturate. An AIS iteration is governed by the following equation, due to Farmer et al [11]:

$$\frac{dx_i}{dt} = c\left[\binom{antibodies}{recognised} - \binom{I\ am}{recognised} + \binom{antigens}{recognised}\right] - \binom{death}{rate}$$

$$= c\left[\sum_{j=1}^{N} m_{ji}x_ix_j - k_1\sum_{j=1}^{N} m_{ij}x_ix_j + \sum_{j=1}^{n} m_{ji}x_iy_j\right] - k_2x_i \quad (1)$$

Where:

$N$ is the number of antibodies and $n$ is the number of antigens.

$x_i$ (or $x_j$) is the concentration of antibody $i$ (or $j$)

$y_j$ is the concentration of antigen $j$

$c$ is a rate constant

$k_1$ is a suppressive effect and $k_2$ is the death rate

$m_{ji}$ is the matching function between antibody $i$ and antibody (or antigen) $j$

As can be seen from the above equation, the nature of an idiotypic interaction can be either positive or negative. Moreover, if the matching function is symmetric, then the balance between "I am recognised" and "Antibodies recognised" (parameters $c$ and $k_1$ in the equation) wholly determines whether the idiotypic effect is positive or negative, and we can simplify the equation. We can simplify the equation still further if we only allow one antigen in the AIS. The simplified equation looks like this:

$$\frac{dx_i}{dt} = k_1 m_i x_i y - \frac{k_2}{n}\sum_{j=1}^{n} m_{ij}x_ix_j - k_3 x_i \quad (2)$$

Where:

$k_1$ is stimulation, $k_2$ suppression and $k_3$ death rate

$m_i$ is the correlation between antibody $i$ and the (sole) antigen

$x_i$ (or $x_j$) is the concentration of antibody $i$ (or $j$)

$y$ is the concentration of the (sole) antigen

$m_{ij}$ is the correlation between antibodies $i$ and $j$

$n$ is the number of antibodies.

In the new equation, the first term is simplified as we only have one antigen, and the suppression term is normalised to allow a 'like for like' comparison between the different rate constants. $k_1$ and $k_2$ were varied as described in section 3. $k_3$ was fixed at 0.1, while the concentration range was set at 0–100 (initially 10). We fixed $n$ at 100. The matching function is the absolute value of the Pearson correlation measure. This allows us to have both positively and negatively correlated users in our neighbourhood, which increases the pool of neighbours available to us.

The AIS is considered stable after iterating for ten iterations without changing in size. Stabilisation thus means that a sufficient number of 'good' neighbours have been identified and therefore a prediction can be made. 'Poor' neighbours would be expected to drop out of the AIS after a few iterations.

Once the AIS has stabilised using the above algorithm, we use the antibody concentration to weigh the neighbours. However, early experiments showed that the most recently added antibodies were at a disadvantage compared to earlier antibodies. This is because they have had no time to mature (i.e. increase in concentration). Likewise, the earliest antibodies had saturated. To overcome this, we reset the concentrations and allow a limited run of the AIS to differentiate the concentrations:

*Reset AIS (set all antibodies to initial concentrations)*

*WHILE (No antibody at maximum concentration) DO*

    *Iterate AIS*

OD

## 2.5 PREDICTION

We predict a rating $p_i$ by using a weighted average over $N$, the neighbourhood of $u$, which was taken as the entire AIS.

$$p_i = \bar{u} + \frac{\sum_{v \in N} w_{uv}(v_i - \bar{v})}{\sum_{v \in N} w_{uv}} \quad (4)$$

$$w_{uv} = r_{uv} x_v \quad (NB\ relative\ not\ absolute)$$

Where $w_{uv}$ is the weight between users $u$ and $v$, $r_{uv}$ is the correlation score between $u$ and $v$, and $x_v$ is the concentration of the antibody corresponding to user $v$.

## 2.6 EVALUATION

- Prediction Accuracy: We take the mean absolute error, where $n_p$ is the number of predictions:

$$MAE = \frac{\sum |actual - predicted|}{n_p} \quad (5)$$

- Mean number of recommendations: This is the total number of unique films rated by the neighbours.

- Mean overlap size: This is the number of recommendations that the user has also seen.

- Mean accuracy of recommendations: Each overlapped film has an actual vote (from the antigen) and a predicted vote (from the neighbours). The overlapped films were ranked on both actual and predicted vote, breaking ties by movie ID. The two ranked lists were compared using Kendall's Tau $\tau$. This measure reflects the level of concordance in the lists by counting the number of discordant pairs. To do this we order the films by vote and apply the following formulae:

$$\tau = 1 - \frac{4 N_D}{n(n-1)}$$

$$N_D = \sum_{i=1}^{n} \sum_{j=i+1}^{n} D(r_i, r_j) \quad (6)$$

$$D(r_i, r_j) = \begin{cases} 1\ if\ r_i > r_j \\ 0\ otherwise \end{cases}$$

Where $n$ is the overlap size and $r_i$ is the rank of film $i$ as recommended by the neighbourhood. Note that i here refers to the antigen rank of the film, not the film ID. $N_D$ is the number of discordant pairs, or, equivalently, the expected cost of a bubble sort to reconcile the two lists. $D$ is set to one if the rankings are discordant.

- Mean number of reviewers. This is the number of reviewers looked at before the AIS stabilised.

- Mean number of neighbours: This is the final number of neighbours in the stabilised AIS.

## 3 EXPERIMENTS

Experiments were carried out on a Pentium 700 with 256MB RAM, running Windows 2000. The AIS was coded in Java[TM] JDK1.3. Each run involved looking at up to 15,000 reviewers (a random sample of up to 20% of the EachMovie data set) to provide predictions and recommendations for 100 users. Averaged statistics are then taken for each run. Runtimes ranged from 5 to 60 minutes, largely dependent on the number of reviewers.

### 3.1 EXPERIMENTS ON SIMPLE AIS

Initial experiments concentrated on a simple AIS, with no idiotypic effects. The goal was to find a good stimulation rate, but also to ensure that the 'baseline' system operates similarly to a SP predictor. Therefore, we set the suppression rate to zero, and varied only the stimulation rate, i.e. The weighting given to antigen binding. Other parameters had been fixed by preliminary experiments to values that worked well with both AIS and SP..

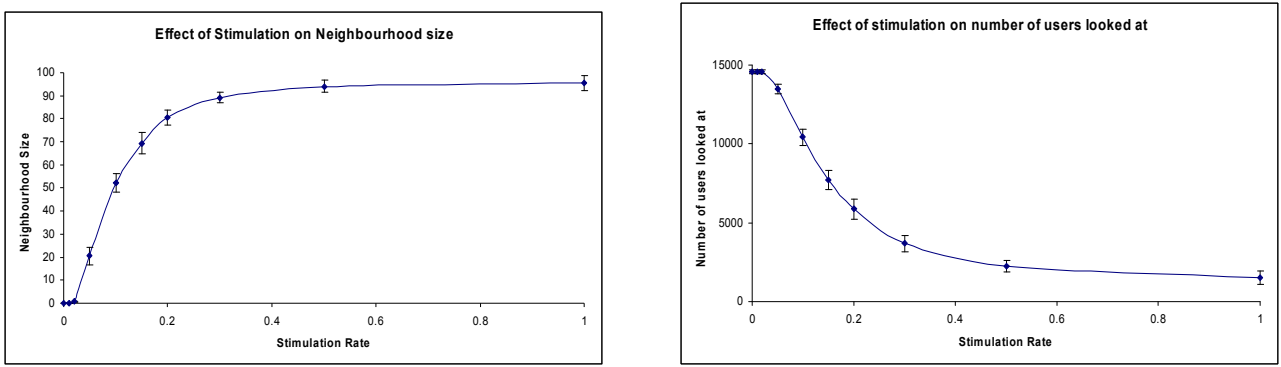

Figure 1: Effect of stimulation rate on neighbourhood and reviewers.

The graphs show averaged results over five runs at each stimulation rate. The bars show standard deviations. In order to have a fair comparison, the SP parameters (neighbourhood and number of reviewers looked at) match the AIS values for each rate. In figure 2, we show the prediction error, number of recommendations, number of overlaps and recommendation accuracy for each algorithm. Note that low prediction error values are better, whereas for the other measures we are looking for high values.

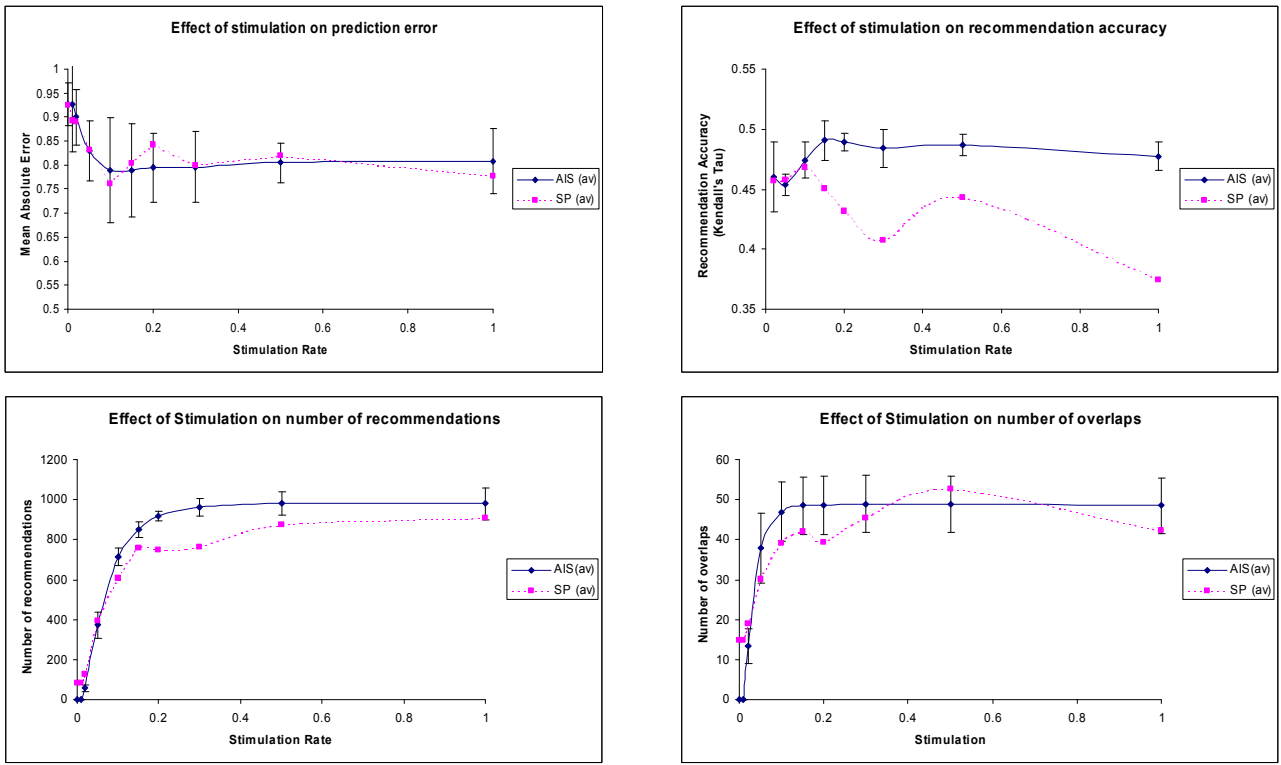

Figure 2: Effect of stimulation rate on prediction and recommendation.

It can be seen that the simple AIS gives broadly similar prediction performance to the SP. The Mean Absolute Error (MAE) measurements from different runs are not normally distributed, so a non-parametric statistic is appropriate. We performed a Wilcoxon analysis, which showed no significant difference between prediction errors of SP and AIS (at a 95% confidence level). In addition, the choice of an appropriate stimulation rate did make a significant difference (comparing a rate of 0.2 with 0.02 at the 95% level).

For recommendation, the AIS performs better than the SP at stimulation rates above 0.1. Again, we performed a positive 95% Wilcoxon analysis to assess significance. We excluded cases where a recommendation score was unavailable (due to an insufficient number of overlaps). The number of recommendations and overlaps show similar trends though the AIS gives a more constant value. Again, some stimulation was beneficial.

In later experiments, the stimulation rate was fixed at one of the better values (0.2, 0.3 or 0.5), in order to give us a good base to work on. These values give us generally good performance, while keeping a good neighbourhood size and still evaluating a reasonable number of reviewers.

## 3.2 EXPERIMENTS ON THE IDIOTYPIC AIS

Having fixed all the simple parameters, we tested the effect of suppression for stimulation rates of 0.2, 0.3 and 0.5. Not surprisingly we found that suppression changed the number of reviewers looked at and the number of neighbours (figure 3):

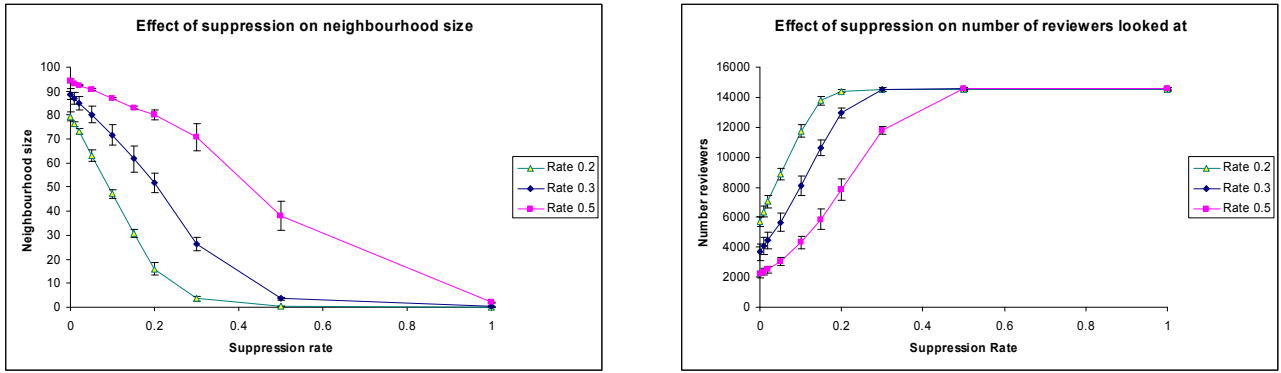

Figure 3: Effect of suppression rate on neighbourhood size and reviewers.

We then tested the effect of suppression on the AIS performance. Here we fixed the baseline rate at stimulation only (no suppression), and took measurements relative to this baseline (Figure 4). Again, it should be noted that the first graph shows prediction error (hence, a good result is low).

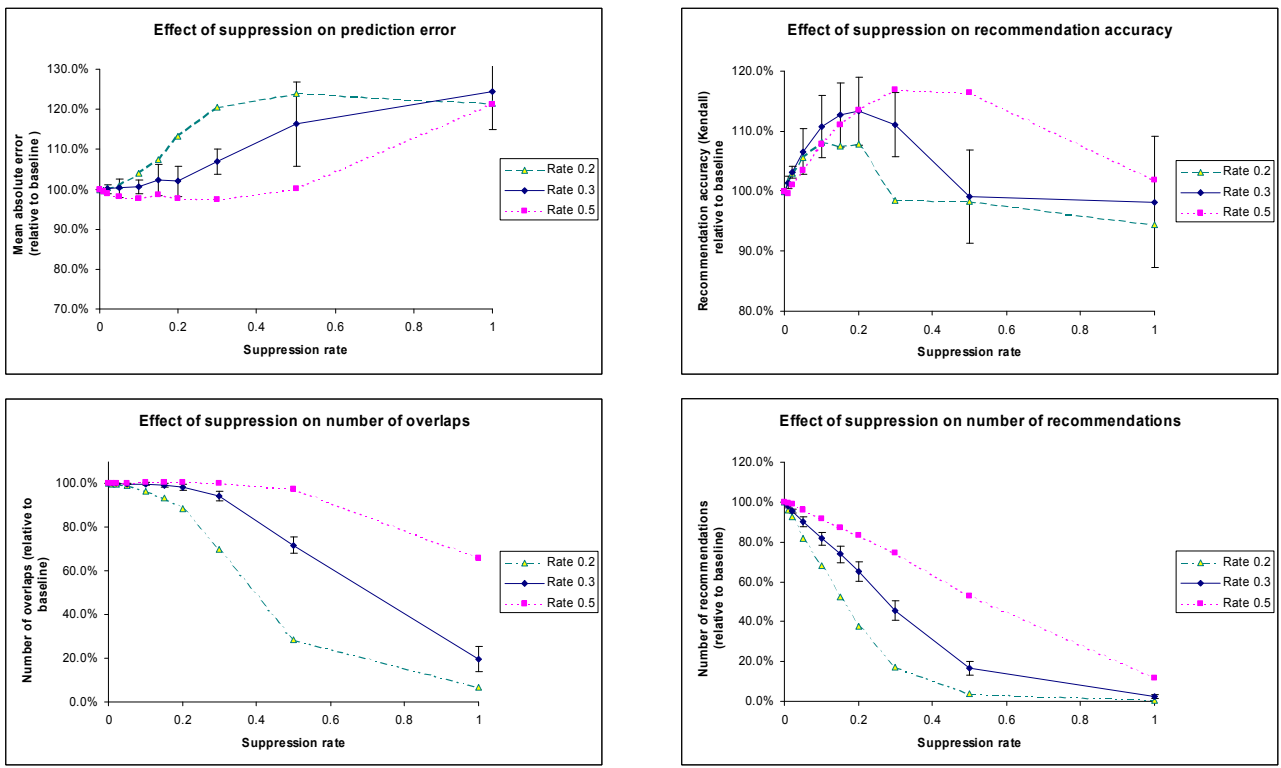

Figure 4: Effect of suppression rate on prediction and recommendation.

Again, the graphs show averaged results over five runs at each suppression rate. The bars show standard deviations (similar size bars for rates 0.2 and 0.5 have been omitted in the interests of clarity). At low levels of stimulation, prediction accuracy is not significantly affected. However recommendation accuracy is improved significantly (95% Wilcoxon). For instance, for 0.3 stimulation, rates from 0.05 to 0.2 gave a significantly improved performance. In actual terms, the Kendall measure rises from 0.5 to nearly 0.6. This means that the chance of any two randomly sampled pairs being correctly ranked has risen from 60% to 80%. Too much suppression had a detrimental effect on all measures.

# 4 IDIOTYPIC ANALYSIS

## 4.1 INTRODUCTION

It has previously been shown that a recommender based on immune system idiotypic principles can outperform one based on correlation alone. However, so far we have not explored the mechanisms of that beneficial effect. Such an exploration would seem worthwhile, particularly if this results in identifying the underlying causes of the improvements of the 'characteristics' of a community (either by changing its membership, or by evaluating the relative merit of each member). Such an effect will be generally useful in a range of applications, of which recommender systems provide just one example. In addition, a deeper understanding of the idiotypic effect may prove useful to the designers of other AIS applications.

## 4.2 ANALYSIS OF EFFECTS

To compare the two predictors regarding their neighbourhood composition, a test user is taken from a database, and then predictions and recommendations are made for that user. Both predictors work by finding a neighbourhood and using that neighbourhood to produce predictions and recommendations. Although both the AIS and SP recommender algorithms are based on Pearson correlations, they act differently for a number of reasons:

The choice of neighbours is different. In the SP, the 100 highest correlated users (or all users that show any correlation, if there are less than 100) are chosen to form a neighbourhood. In the AIS, this general rule is followed, except that stimulation adds threshold and idiotypic effect adds diversity.

Even given the same neighbours, the weighting is different. In the SP, the neighbour weight is the correlation between that neighbour and the test user. In the AIS, this correlation is multiplied by that antibody's (neighbour's) concentration, which in turn is determined by running the AIS algorithm over the neighbourhood.

To deal with the first point, the stimulation rate provides some fixed threshold for the correlation of any antibody with the antigen. Even in the absence of any idiotypic interactions, an antibody's correlation (weighted by the stimulation rate) must outweigh the death rate; otherwise, it will not survive in the AIS. So, at low stimulation rates it may prove difficult to fill the AIS completely. Conversely, at very high stimulation rates it may not be necessary to examine all the supplied users in order to fill an AIS.

This effect can be seen in Figure 1. Such a thresholding effect has been shown to be beneficial by Gokhale [13] in maintaining the quality of a neighbourhood by filtering out poorly correlated users (the SP will consider all reviewers who have at least one vote in common with the test user). Thus, the idiotypic effect should be viewed in the context of providing further refinement to a neighbourhood that is already known to be in some sense 'good'. Since the effect (in our model) is always negative, its impact may be to improve diversity by removing 'suboptimal' users from the AIS. Conversely, it might be that the idiotypic effect is effective because, given a neighbourhood, it changes the weight of each neighbour (or concentration of each antibody) in that neighbourhood. This is the second point highlighted above.

In order to test out these hypotheses, we took a sample result, based on 100 predictions for detailed analysis. The three settings for each algorithm were as detailed in section 2, except that default votes were not used. Thus, if a neighbour has not seen a film then that neighbour is ignored when making a prediction for that film. The AIS parameters were set to 'good' values (as observed previously). Thus stimulation rate was set to 0.3 and suppression rate to 0.2. As reported previously, the prediction performance (mean absolute error) was not significantly different between the two algorithms, but recommendation (Kendall's Tau) was significantly better for the AIS recommender (as before, a Wilcoxon matched pairs signed rank test was used to assess significance).

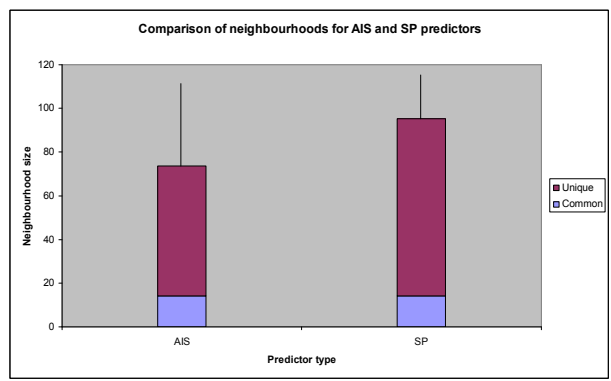

Figure 5: Comparison of AIS and SP neighbourhoods. The total size of each bar represents the total size of the neighbourhoods produced by each predictor (averaged over 100 predictions; bar shows standard deviation). The lower part of each bar shows the average number of common neighbours (i.e. Appearing in both neighbourhoods). The remainder of the bar is composed of unique neighbours – that is, neighbours who appeared in one neighbourhood but not the other.

The first thing to observe is that the neighbourhoods produced by each algorithm are different. As implied from the above, SP tended to produce large neighbourhoods (average 95.4 as opposed to 73.8 using the AIS) and Figure 5 shows that the composition of these neighbourhoods is different. In particular, it does not seem that the AIS neighbourhoods are merely subsets of the SP neighbourhoods. In fact, the vast majority of neighbours are 'unique' – that is, chosen by one algorithm but not the other. Is it the neighbourhoods that make the difference to prediction and recommendation performance? Figure 6 shows AIS and SP performance on both neighbourhoods. For this experiment, we recorded the neighbourhoods found by both the AIS and SP algorithms.

We then reran the predictions, with everything the same except that this time we forced the AIS and SP algorithms to use our 'fixed' neighbourhoods. We can see that for prediction, changing the neighbourhood (or indeed algorithm) did not seem to make any significant difference (Table 1 has the details of the statistical tests). However, for recommendation, although the means are very similar (Figure 6), the AIS neighbourhood usually produced better recommendations than the SP neighbourhood (Table 1b). In fact, the neighbourhood effect seems to dominate, since given the AIS neighbourhood, the SP algorithm appears to do significantly better than the AIS algorithm for recommendation. There is one exception to this, where the AIS algorithm does not do significantly better for either neighbourhood. In addition, the AIS algorithm does better on the SP neighbourhood than the SP algorithm, indicating that the neighbour weightings, as well as the neighbours themselves, also contribute to the recommendation quality.

| 1st Predictor | 1st NH | 2nd Predictor | 2nd NH | Median 1 | Median 2 | Number of comparisons | 1st better | 2nd better | Significance (upper bound) |
|---|---|---|---|---|---|---|---|---|---|
| SP | SP | AIS | SP | 0.682 | 0.697 | 97 | 2212 | 2541 | 0.5551 |
| SP | SP | SP | AIS | 0.682 | 0.658 | 97 | 2163 | 2590 | 0.4434 |
| SP | SP | AIS | AIS | 0.682 | 0.652 | 97 | 2176 | 2577 | 0.4717 |
| AIS | SP | SP | AIS | 0.697 | 0.658 | 97 | 2256 | 2497 | 0.6659 |
| AIS | SP | AIS | AIS | 0.697 | 0.652 | 97 | 2258 | 2495 | 0.6711 |
| SP | AIS | AIS | AIS | 0.658 | 0.652 | 84 | 1706 | 1864 | 0.7263 |
| Results for Predictions above; Result for Recommendations below | | | | | | | | | |
| SP | SP | **AIS** | **SP** | 0.525 | 0.557 | 83 | 801 | 2685 | 1.917e-05 |
| SP | SP | **SP** | **AIS** | 0.525 | 0.549 | 83 | 707.5 | 2778.5 | 2.617e-06 |
| SP | SP | **AIS** | **AIS** | 0.525 | 0.542 | 85 | 930 | 2725 | 8.483e-05 |
| AIS | SP | **SP** | **AIS** | 0.557 | 0.549 | 82 | 1218.5 | 2184.5 | 0.02571 |
| AIS | SP | AIS | AIS | 0.557 | 0.542 | 80 | 1426 | 1814 | 0.3534 |
| **SP** | **AIS** | AIS | AIS | 0.549 | 0.542 | 78 | 2149 | 932 | 0.002459 |

Table 1: Analysis of differences between neighbourhoods (NH) and algorithms for both prediction and recommendation. In each case, the Wilcoxon significance test was applied to the results obtained from each pair of regimes. Regimes that are significantly better are shown in bold (there were no significant differences found for prediction).

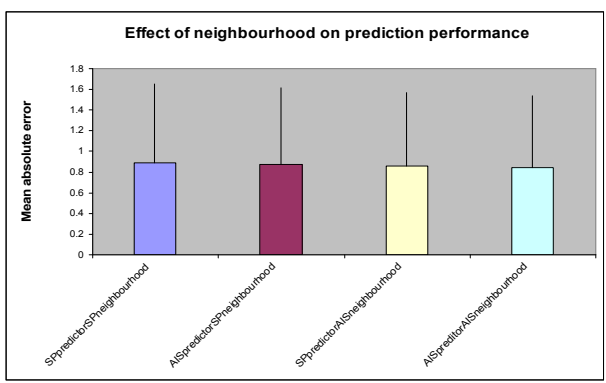 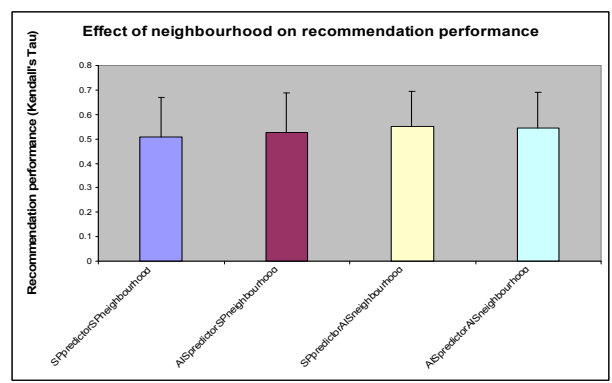

Figure 6: Effect of neighbourhood composition for AIS and SP algorithms. The left graph shows prediction performance (measured as mean absolute error averaged over 100 predictions) for each algorithm and each neighbourhood. The right graph shows recommendation performance deviation (measured as Kendall's Tau averaged over 100 predictions) for each algorithm and each neighbourhood. Bars show standard deviation.

We ran these experiments using default votes (neighbours who had not voted on a film were assumed to give the film a slightly negative rating) and obtained similar results. It is worth pointing out at this stage that these results should not be taken to be exhaustive, merely indicative. Indeed, we would not want to draw any firm conclusions based on only 100 predictions. This point will be returned to later. Nevertheless, the results obtained so far seemed to indicate that it was worth investigating the contribution of neighbourhood composition to recommendation performance.

We looked at a variety of neighbourhood parameters (we might term these community characteristics) across SP and AIS neighbourhoods. Four characteristics are of particular interest, and each will be discussed in turn. Firstly, it might seem reasonable to assume that performance improves with the number of neighbours in a neighbourhood. However, clearly there is a cost in collecting neighbours (of appropriate quality) together, and thus it will be useful if we can provide good quality recommendations from smaller neighbourhoods.

Another characteristic is the overlap size, which governs the number of recommendations we can assess (An overlap is a test user vote that is also contained in the union of all neighbours' votes). Thirdly, we looked at correlation between each neighbour and the test user. A high correlation shows that neighbours are clustered 'tightly' around the test user, which we might imagine would provide for better recommendations. Fourthly, the idiotypic effect is expected to reduce the inter-neighbour correlations. An obvious intuition might be that such a reduction causes an increase in recommendation quality.

Table 2 shows the difference in these community characteristics across SP and AIS neighbourhoods. It can be seen that the AIS does produce neighbourhoods that are measurably different in character to the SP neighbourhoods. In summary, the AIS neighbourhoods are smaller, have less overlap, are generally less correlated with the test user and have lower inter-neighbour correlations. In order to test out which (if any) of these characteristics is crucial, we plotted recommendation performance against each for the AIS algorithm. The results seem to show that none of these characteristics on their own influences the performance in a clear way. Figure 7 shows scatter plots generated for each characteristic against recommendation quality. Trend lines (based on a power law) have been added to emphasise any underlying data trends.

| 1st Predictor | 2nd Predictor | Characteristic tested | Mean 1 | Mean 2 | Unequal neighbour-hoods | 1st has higher value | 2nd has higher value | Significance (upper bound) |
|---|---|---|---|---|---|---|---|---|
| SP | AIS | Neighbours | 95.40 | 73.75 | 97 | 4602 | 151 | 1.196e-15 |
| SP | AIS | Overlap | 47.46 | 46.39 | 26 | 334.5 | 16.5 | 5.686e-05 |
| SP | AIS | Correlation | 0.12 | 0.10 | 79 | 2566 | 594 | 1.465e-06 |
| SP | AIS | Neighbour correlation | 0.15 | 0.04 | 83 | 3477 | 9 | 3.572e-15 |

Table 2: Analysis of difference in neighbourhood characteristics between SP and AIS algorithms. Four characteristics are shown. In each case, the Wilcoxon significance test was applied to the neighbourhoods obtained from the algorithms. In all four cases, the value for the SP was significantly higher; this is indicated by bold type.

The first plot suggests that neighbourhood size is not essential in order to obtain high quality recommendations. The second plot, however, does suggest that small overlap sizes might be beneficial for producing good recommendations (regression analysis has not been performed so at this stage this is merely a suggestion). This in some sense is intuitive, as it might be easier to produce higher quality recommendations if there are less of them. However, a balance needs to be struck here; once the overlap size gets too low, the neighbourhood may no longer prove useful to the user. The third plot shows that, perhaps surprisingly, high correlation between neighbours and the test user may not be essential for high quality recommendations. Finally, the fourth plot would seem to indicate that reduced inter-neighbour correlation is not important in recommendation accuracy, or at least if it is responsible, it is part of a wider effect.

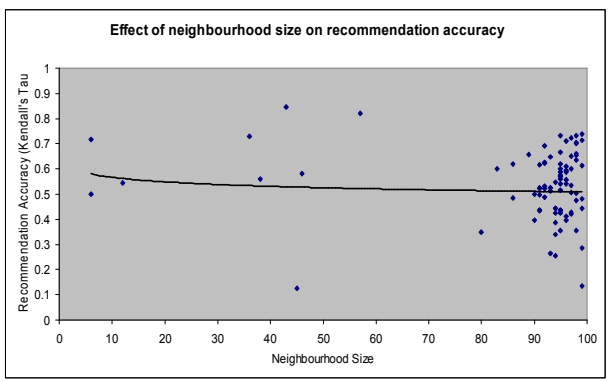 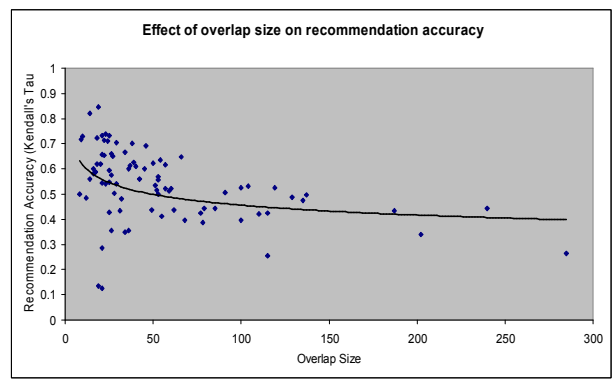

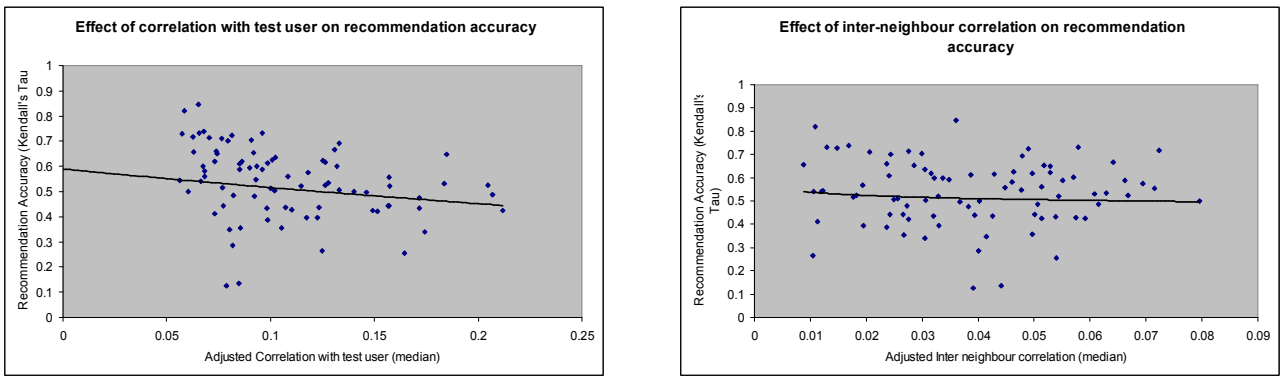

Figure 7. Effect of various neighbourhood measures on AIS recommendation performance. In each graph, the measure is shown on the x-axis. The recommendation performance (where available) for each of 100 AIS predictions is plotted against this neighbourhood measure. Trend lines are added to indicate the underlying data trend (if any).

## 5   DISCUSSION AND CONCLUSIONS

It is not particularly surprising that the simple AIS performs similarly to the SP predictor. This is because they are, at their core, based around the same algorithm. The stimulation rate (in absence of any idiotypic effect) is effectively setting a threshold for correlation. This has both strengths and weaknesses. It has been shown that a threshold is useful in discarding the potentially misleading predictions of poorly correlated reviewers [13]. On the other hand, a rigid threshold means that one has to 'prejudge' the appropriate level to avoid both premature convergence and empty communities. Indeed, detailed examination of the individual runs showed that the AIS had a tendency to fill its neighbourhood either early or not at all. The setting of a threshold also means that sufficiently good antibodies are taken on a first come, first served basis. It is interesting to observe that such a strategy nevertheless seems (in these experiments) to provide a more constant level of overlaps, and better recommendation quality.

The richness of our AIS model comes when we allow interactions between antibodies. Early, qualitative experimentation with the idiotypic network showed antibody concentration rising and falling dynamically as the population varied. For instance, in the simple AIS, the concentration of an antibody will monotonically increase to saturation, or decrease to elimination, unaffected by the other antibodies. However, there is a delicate balance to be struck between stimulation and suppression. An imbalance may lead to a loss in population size or diversity. The graphs show that a small amount of suppression may indeed be beneficial to AIS performance, in particular recommendation. It is interesting to note that the increase in recommendation quality occurs with a relatively constant overlap size. At too high levels of suppression, it is harder to fill the neighbourhood, with consequent lack of diversity and hence recommendation accuracy.

We believe that these initial results show two things. Firstly, population effects can be beneficial for CF algorithms, particularly for recommendation; secondly, that CF is a promising new application area for AISs. In fact, we can widen the context, since the process of neighbourhood selection described in this paper can easily be generalized to the task of ad-hoc community formation. As mentioned previously, it is not claimed that these results are conclusive. Indeed, much more data is required before any firm conclusions can be drawn. In this respect, this paper is very much a work in progress. Nevertheless, the results to date certainly are indicative, and challenge certain assumptions. It is hoped that the presentation of these results will stimulate discussion and interest in the nature of the idiotypic effect.

It does not seem likely that the idiotypic effect can be captured by one particular measurement as it is likely to be a combination of factors. For example, we have shown that both the neighbourhood choice and the weighting of neighbours within that neighbourhood can influence the recommendation performance. There are further community characteristics that could be explored. Some (for example, number of recommendations, overlaps per neighbour, absolute correlation scores) have been examined and shown to be inconclusive. Some (for example, number of neighbours voting on each film) remain potential future subjects for investigation. Other tests (e.g. setting each neighbour's concentration to a random number for immune system predictions, to see whether accurate concentrations are really necessary) might shed further light on the relative importance of each measure.

There are wider implications for such work. The database used for this study [8] is based on real peoples' profiles. Thus, any headway made into improving neighbourhoods by the idiotypic effect can have real benefit for other recommenders – and indeed any community based application. Current research is under way to extend this work to predict websites of interest based on users' bookmarks [20].